\documentclass[10pt,twocolumn,letterpaper]{article}

\usepackage{graphicx}
\usepackage{amsmath}
\usepackage{amssymb}
\usepackage{booktabs}
\usepackage{subcaption}
\usepackage{tcolorbox}

\usepackage[pagebackref,breaklinks,colorlinks]{hyperref}

\usepackage[capitalize]{cleveref}
\crefname{section}{Sec.}{Secs.}
\Crefname{section}{Section}{Sections}
\Crefname{table}{Table}{Tables}
\crefname{table}{Tab.}{Tabs.}

\begin{document}

\title{Deep Generative Neural Embeddings  for \\ High Dimensional Data \\
Visualization }

\date{\makebox{}}
\author{January 25, 2023\\
    Halid Ziya Yerebakan \\
  Gerardo Hermosillo Valadez\\
  Siemens Medical Solutions USA\\
  40 Liberty BLVD, Malvern, PA, 19355 \\
  \texttt{halid.yerebakan@siemens-healthineers.com}\\
  \texttt{gerardo.hermosillovaladez@siemens-healthineers.com}}

\maketitle

\begin{abstract}
    We propose a visualization technique that utilizes neural network embeddings and a generative network to reconstruct original data. This method allows for independent manipulation of individual image embeddings through its non-parametric structure, providing more flexibility than traditional autoencoder approaches. We have evaluated the effectiveness of this technique in data visualization and compared it to t-SNE and VAE methods. Furthermore, we have demonstrated the scalability of our method through visualizations on the ImageNet dataset. Our technique has potential applications in human-in-the-loop training, as it allows for independent editing of embedding locations without affecting the optimization process.
\end{abstract}

\section{Introduction}
\label{sec:intro}


Visualization plays a crucial role in understanding and interpreting complex data sets. It is an essential tool for evaluating the performance of feature representations, identifying patterns and sub-groups within data, detecting errors in data selection, and comparing the difficulty of decision boundaries across different classes. However, high-dimensional data visualization is a challenging task due to the curse of dimensionality. To overcome this challenge, researchers in the machine learning community continue to explore and develop new visualization techniques to handle high-dimensional data effectively.

Linear projection methods are not sufficient to capture non-linear manifolds in contemporary high-dimensional datasets. Thus, TSNE and UMAP are the common choices for high-dimensional data visualization in recent machine learning literature \cite{mcinnes2018umap,van2008visualizing}. They create a neighborhood graph of data in original dimensions and preserve these graph relationships in lower-dimensional space by gradient-based optimization methods. This allows for a more accurate representation of the underlying structure of the data.


Auto Encoder (AE) is a complete parametric method that could provide visualizations if the bottleneck dimensions are restricted to two dimensions \cite{Kingma2014AutoEncodingVB}. It consists of two networks: an encoder network that maps the input data into a latent space that could be used for visualization and a decoder network that maps points from the latent space back to the original input space. The visualization layer is a function of the encoder portion of the network, thereby preventing any optimization for visualization independent of individual data points. Also, two dimensions in bottleneck are very restrictive for effective information flow in most cases.

In this paper, we propose a practical method to create a visualization while obtaining a generative model of the data without an encoder similar to Park et. al. \cite{park2019deepsdf}. To achieve this, we relaxed the encoder part of Variational Auto Encoder (VAE) and let it be learned as embeddings. We name this class of models as Generative Neural Embeddings (GNE). The optimization objective maximizes the embedding locations and generator (decoder) network jointly. We compared GNE generated visualizations versus TSNE and variational auto encoder approaches. Additionally, we demonstrated the ability to generate new samples from embeddings. Finally, we have demonstrated the scalability of our method on the Imagenet dataset while comparing against VAE.

\section{Method}

	Our method could be described as an embedding layer and a generative model on this embedding layer. Embedding layer $E$ provides the lookup table of embeddings for data point id $i$. Generative model $G$ generates the output from the given embedding. Thus, the generation process for data $i$ becomes $G(E[i])$. The objective of the whole network is to try to minimize the loss between generated data versus the real data corresponding to id $i$. Thus it is a compression of whole data into a single network. The purpose of the embedding is to make the inputs of the generator optimizable. Hence, the name generative network embeddings (GNE) comes from this connection. Since the number of parameters grows with the number of data points, this method is a non-parametric method. 
	
	 Any generative model is applicable to our method. We have selected a fully connected resnet structure for the generative part of the model \cite{he2016deep} in our first set of experiments which shows a basic construction. In order to have enough number of hidden units in the network width, there is a dense expansion layer that increases the number of dimensions after the embeddings. Embeddings are selected to be in 2 dimensions to project into a plane. However, any number of dimensions is possible for other purposes. In our model, we have an additional Gaussian noise layer that regularizes the embedding space. An example Keras implementation of the model is shown in the verbatim listing where $NW, EPS$ and $NLAYERS$ are network hyperparameters.
	 

    








%


\begin{tcolorbox}
\begin{verbatim}
dims = data.shape # Nxd
lin = Input(1)
embed = Embedding(dims[0], 2)(lin)
flatten = Flatten()(embed)
noise = GaussianNoise(EPS)(flatten)
hidden = Dense(NW, 'tanh')(noise)
for l in range(NLAYERS):
    relu = Dense(NW, 'relu')(hidden)
    linear = Dense(NW)(relu)
    hidden = add([hidden, linear])
out = Dense(dims[1], 'sigmoid')(hidden)
\end{verbatim}
\end{tcolorbox}

\section{Results and Discussion}

\begin{figure*}
  \centering
  \begin{subfigure}{0.48\textwidth}
    \includegraphics[width=\textwidth]{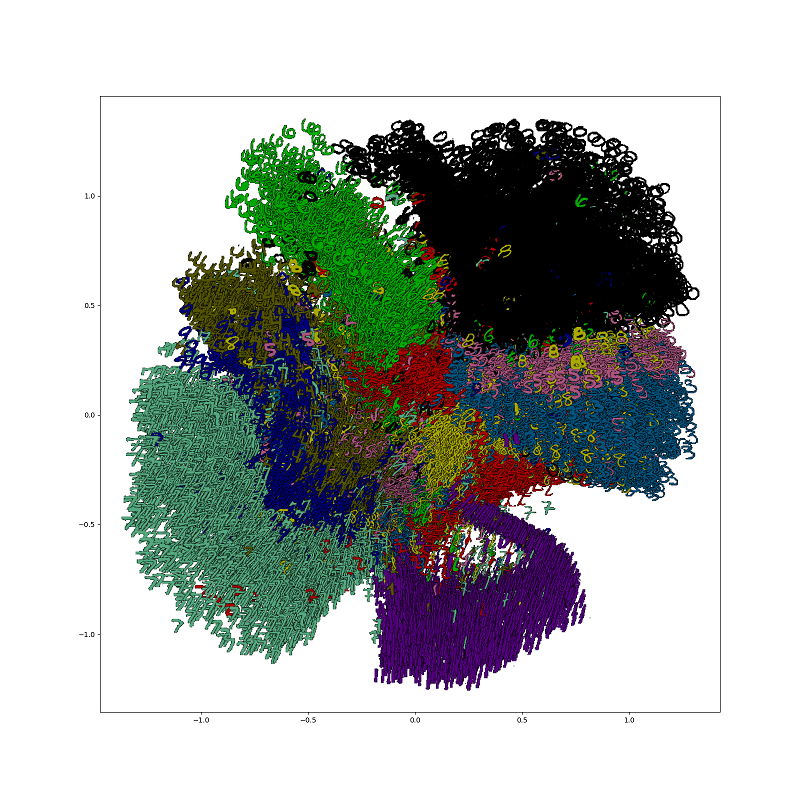}
    \caption{2 dimensional scatter plot of GNE embeddings}
    \label{fig:gne}
  \end{subfigure}
  \hfill
  \begin{subfigure}{0.44\textwidth}
	\centering
    \includegraphics[width=\textwidth]{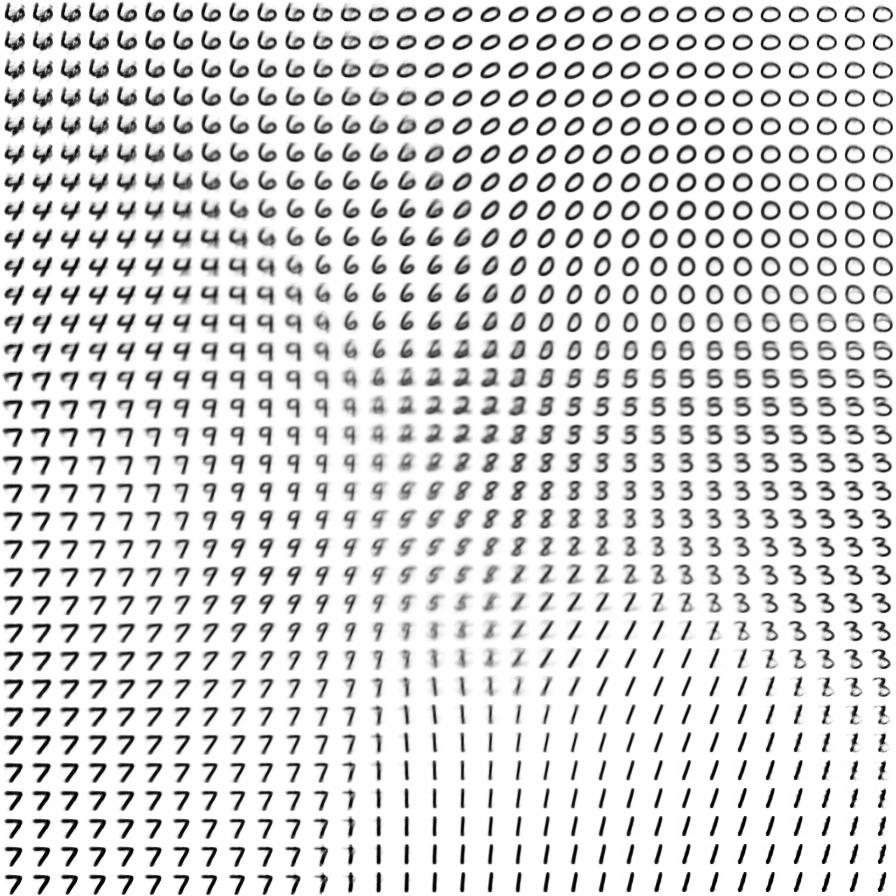}
    \caption{Generated Digits from GNE decoder}
    \label{fig:generation}
  \end{subfigure}
  \caption{Generative Network Embeddings}
  \label{fig:gne}
  
   \vskip\baselineskip
   
 \centering
  \begin{subfigure}{0.48\textwidth}
    \includegraphics[width=\textwidth]{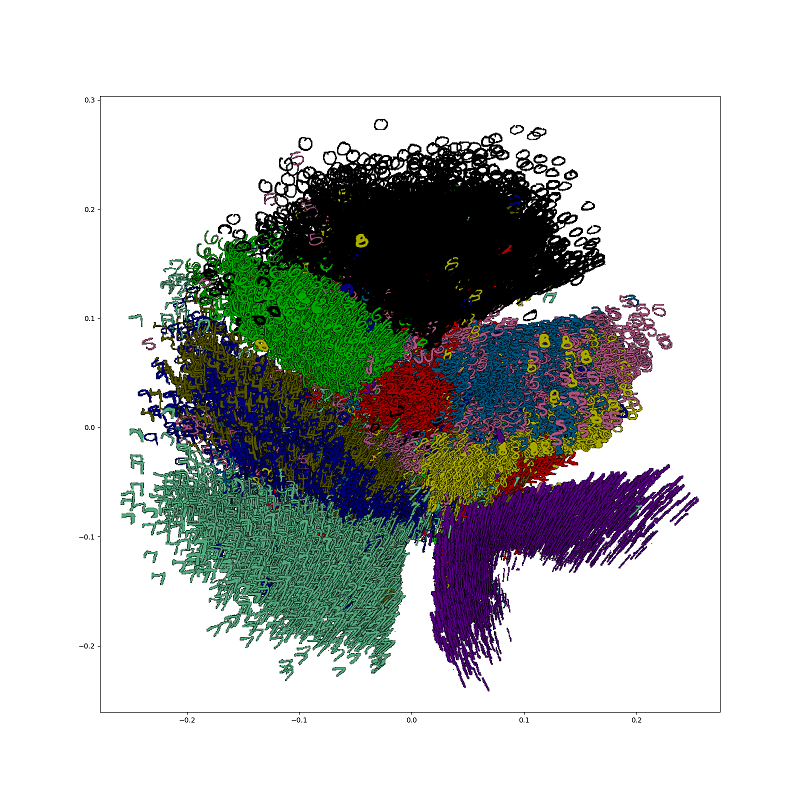}
    \caption{2 dimensional scatter plot of VAE encodings}
    \label{fig:vae}
  \end{subfigure}
  \hfill
  \begin{subfigure}{0.44\textwidth}
	\centering
    \includegraphics[width=\textwidth]{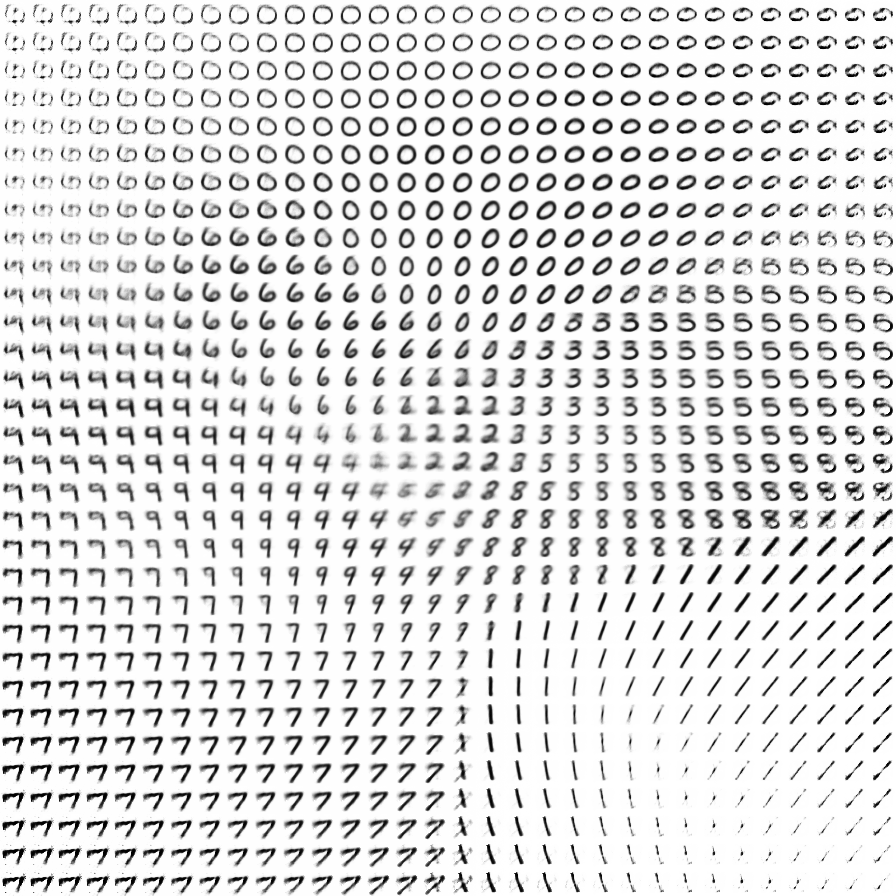}
    \caption{Generated Digits from VAE decoder}
   \label{fig:vae_generation}
  \end{subfigure}
  \caption{Variational Auto Encoder}
  \label{fig:vae}

\end{figure*}

\begin{figure*}

  \vskip\baselineskip
    \centering
    \begin{subfigure}{0.48\textwidth}
    \includegraphics[width=\textwidth]{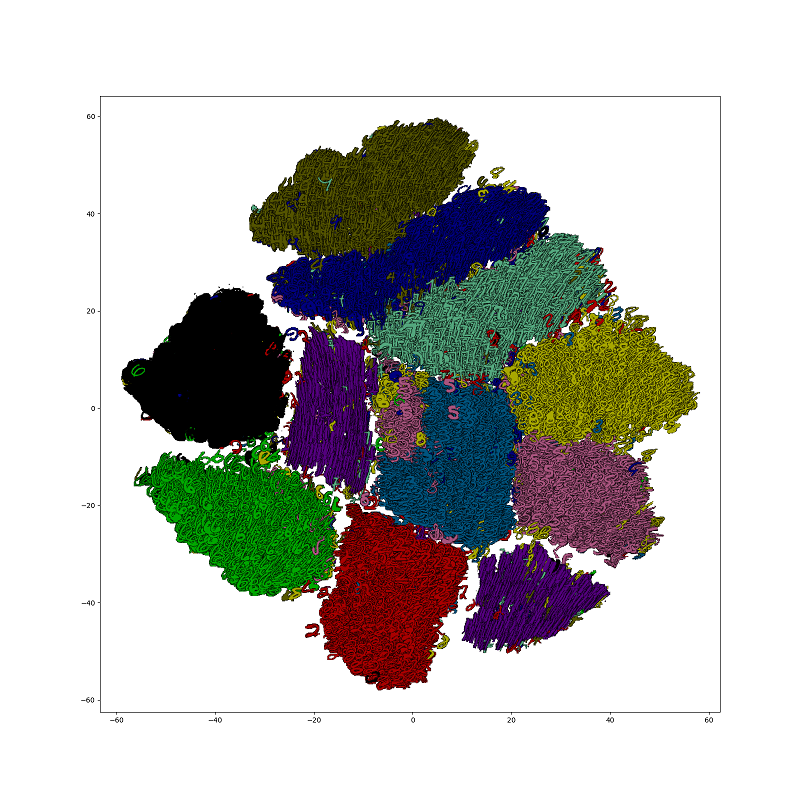}
    \end{subfigure}
  \caption{2 dimensional scatter plot of TSNE embeddings}
 \label{fig:tsne}

\end{figure*}

In the first experiment, we used the MNIST training dataset after normalizing grayscale intensities to 0-1 scale. The embedding size becomes 60000x2 for this dataset. We used Adam optimizer for training with an initial learning rate of 1e-3 with a batch size of 1024. We selected 64 dimensions as the number of hidden dimensions for 4 layers of residual blocks. We have trained models for 100 epochs. We used mean squared error as the loss function. We have selected 1e-2 for coefficients in the noise term of VAE and 1e-1 for GNE. Higher value coefficients deteriorate reconstruction due to the bottleneck of 2 dimensions. 

Figures \ref{fig:gne} and \ref{fig:vae} show the visualizations of the dataset for GNE and VAE, respectively. In the scatter plots, each color represents a distinct digit class, and the coordinates are embedding values for the corresponding digit image. Additionally, we have demonstrated the generation ability of both decoders in grid points corresponding to the visualization dimension by generating images on those locations. For reference, TSNE embeddings on the dataset are shown in Figure \ref{fig:tsne}. Both neural network visualizations exhibit more mixes of the classes on the edges since they do not use pairwise similarity information as TSNE does. Both neural networks exhibit similar behavior in transitions for the grid visualizations, while GNE is about twice as fast (1s 13ms per epoch vs. 2s 35us). Final loss values are comparable and change depending on Gaussian noise scale parameters in both networks.

\begin{figure*}
  \centering
    \begin{subfigure}{0.99\textwidth}
    \includegraphics[width=\textwidth]{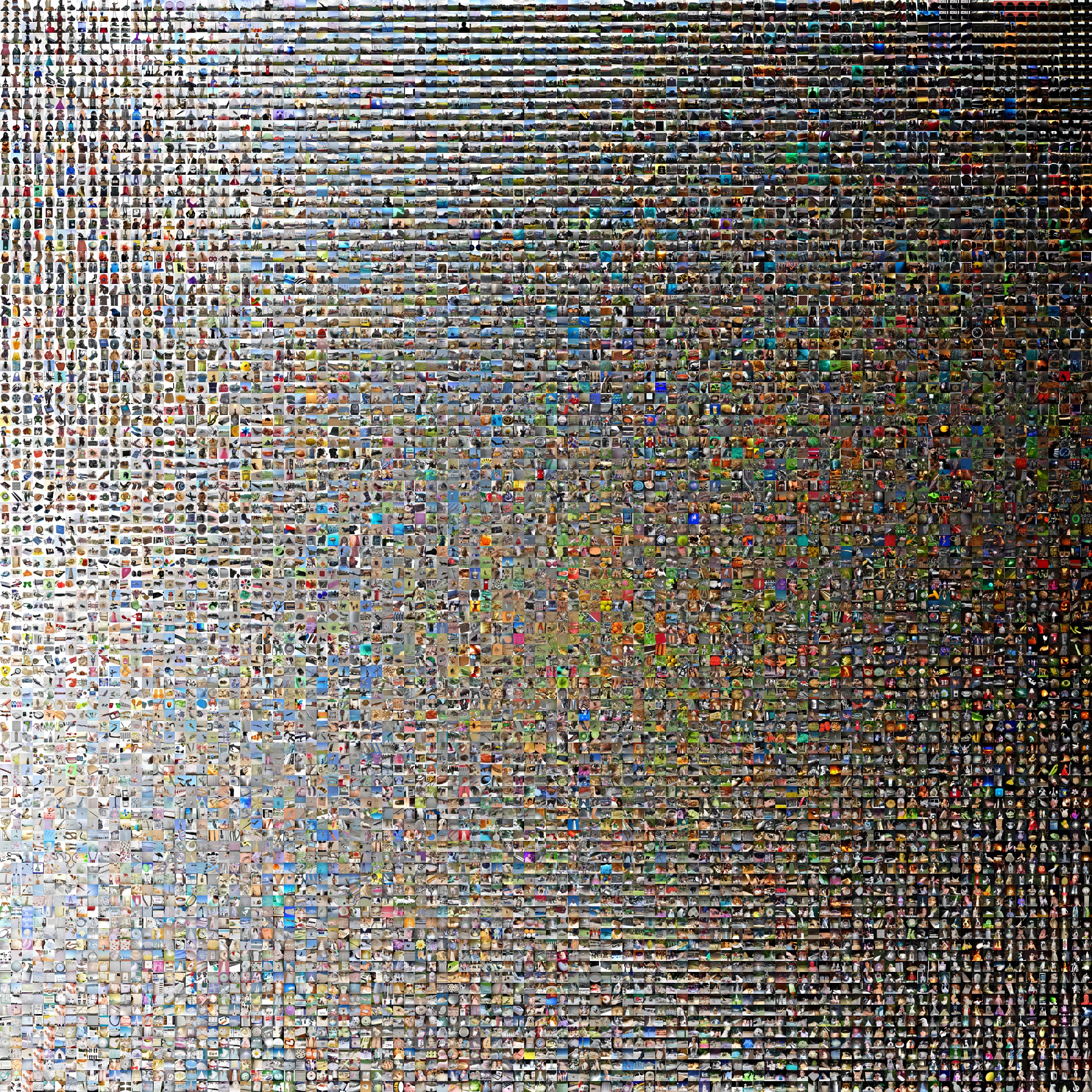}
    \end{subfigure}
    \caption{Visualization of nearest neighbors on grid points respect to VAE encodings}
    \label{fig:imagenetvae}
\end{figure*}

\begin{figure*}
  \centering
  \begin{subfigure}{0.99\textwidth}
    \includegraphics[width=\textwidth]{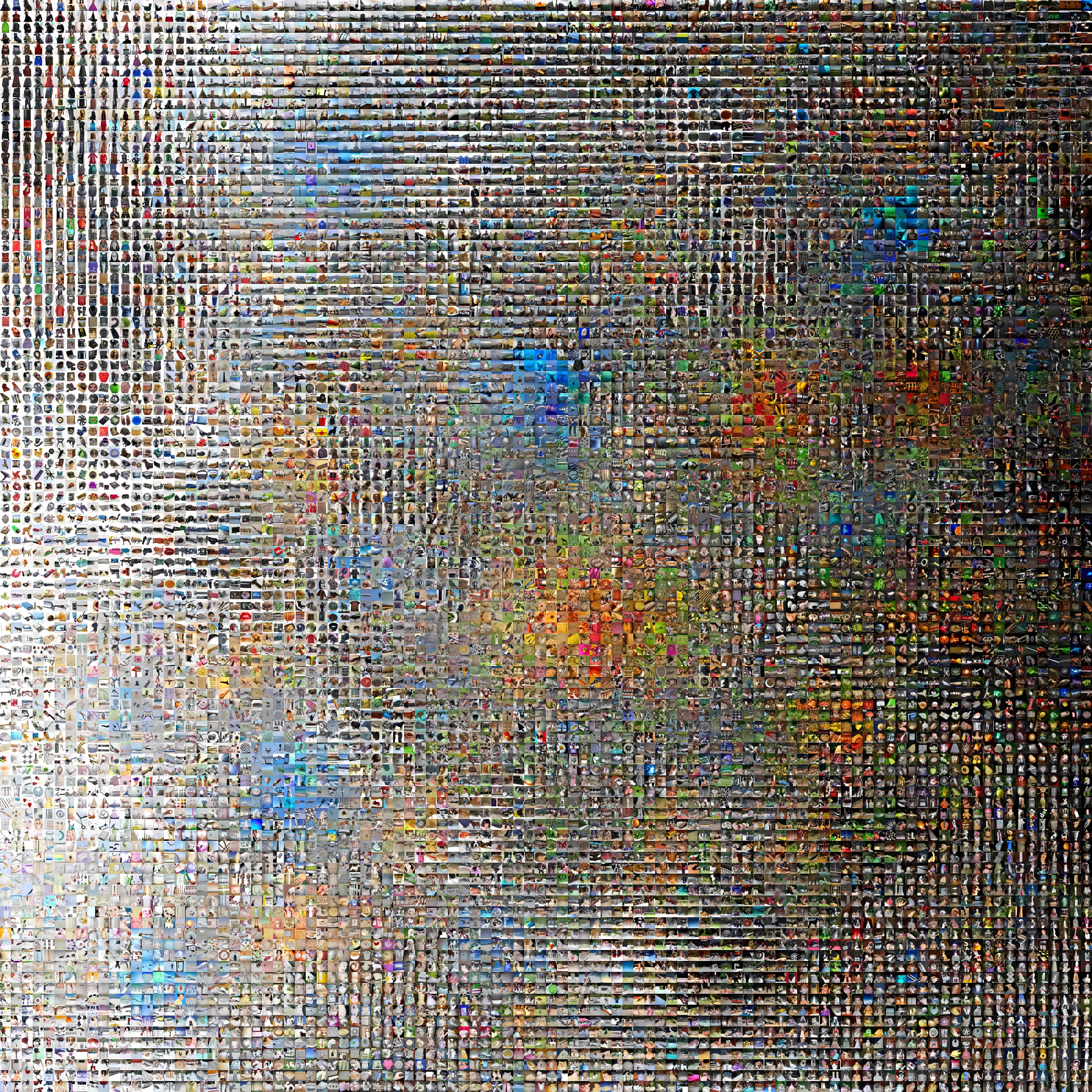}
  \end{subfigure}
    \caption{Visualization of nearest neighbors on grid points respect to GNE embeddings}
    \label{fig:imagenetgne}
\end{figure*}

For large-scale experimentation, we have used Imagenet ILSVRC 2017 training subset with 1,281,167 images. We have reduced the resolution to 28x28 after cropping images from their center. In this experiment, we have used a deconvolutional generator with 128 dimensions for VAE and GNE. VAE gives 0.379 mean squared error and GNE gives 0.376 with convergence. GNE takes 130s per epoch, whereas VAE takes 446s per epoch. Due to the flexibility of GNE, we could use VAE encodings as initialization and continue training. This setting achieved 0.322 mse. Qualitatively, GNE improves the local contrast after the initialization with VAE estimations as shown in Figures \ref{fig:imagenetvae} and  \ref{fig:imagenetgne}. In these figures, we have used the real nearest neighbors instead of the generated images. Note that visualizations are generated for raw data, not for semantic features that is extracted from some other network.


\section{Related Work}



Word2vec \cite{nipsword2vec,Mikolov2013EfficientEO} uses a similar idea on the generation of textual data by hierarchical softmax given the word embeddings. In our case, we have used a deep residual network for image generation and gave dummy ids to each image data point. Additionally, low-dimensional embeddings are chosen for visualization purposes.

DeepDream is a computer vision algorithm that uses a deep neural network to enhance patterns in images \cite{deepdream}. DeepDream optimizes the input image to maximize the activation of the selected neuron with static network parameters. In our case, we are optimizing input embeddings of the generative model while allowing optimization on the generative model. 

Our work is related to Generative Adversarial Networks  \cite{NIPS2014_5ca3e9b1} in terms of image generation. The networks are optimized in pairs as discriminator and generator in these methods. The generator part tries to map random noise into realistic data such that the discriminator would fail to discriminate between synthetic and real data. In our method, however, the generator tries to reconstruct the original images themselves. Also, embeddings are optimized instead of input sampling from a random distribution. 

Park et al. introduced DeepSDF, a non-parametric auto decoder network, in a previous study \cite{park2019deepsdf}. Similarly to our method, their method relaxes the encoder part of the auto encoder networks. In their paper, they have focused on shape generation applications. In our study, we investigated the utilization of similar architecture for visualization, which creates the opportunity to broaden aspects of auto decoders. Also, we do not require any prior distribution assumption on embeddings.

\section{Future Works}

	We have demonstrated the utilization of embeddings to generate visualizations of data along with a generative model. These kinds of models have more potential than what we described in this paper. Thus, we would like to list a few of them in this section. 
	
	First, the method could model each class using multi-modal distributions, and there is no explicit constraint that enforces similar points to appear in a similar location in the visualization space. On the other hand, adding pairwise similarity constraints like in neighborhood graph models increases the expected computational time. Thus investigating this tradeoff could be an interesting research direction.   

	Second, the optimization algorithm is gradient-based in this study. It does not have to be since the input space is very small. For example, even a global optimization with a simple grid search in 2D space could yield a better grouping of data points. However, this optimization splits the optimization of the decoder part of the network from embeddings. 
	
	Third, GNE does not describe a direct way for getting embeddings for test data in the case of a train-test split. One way to obtain them is by optimizing the embedding part of test cases while keeping the generator fixed like DeepDream\cite{deepdream}. As expected, the optimization approach would be slower than standard feed-forward architectures in runtime.
	
	Fourth, thanks to independent embeddings, individual elements in the visualization could be modified with human interaction. For example, a user could select a set of embeddings and drags them to a place that he thinks has more similar data points. This manipulation improves to chance of getting better optimization results while allowing interactive exploration. Additionally, decision boundaries could be mapped back to this visualization space for manual edits. The effect of this visualization method on data labeling speed is worth investigating. 
	
	
	

{\small
\bibliographystyle{ieee_fullname}
\bibliography{gne}
}

\end{document}